\documentclass[3p,times]{elsarticle}
\usepackage{hyperref}

\journal{Journal of \LaTeX\ Templates}

\usepackage{amsmath,amssymb,amsthm}
\usepackage{algorithm,algorithmic} 
\usepackage{booktabs}
\usepackage{subfigure}
\usepackage[retainorgcmds]{IEEEtrantools}
\usepackage{hyperref}

\newtheorem{example}{E{\scriptsize  xample}}[section]

\def\R{\mathbb{R}}

\def\H{\mathbb{H}}

\newcommand{\iq}{{\bf i}}
\newcommand{\Iq}{{\bf I}}
\newcommand{\jq}{{\bf j}}
\newcommand{\kq}{{\bf k}}

\newcommand{\aq}{{\bf a}}

\newcommand{\Vq}{{\bf V}}
\newcommand{\vq}{{\bf v}}
\newcommand{\Uq}{{\bf U}}
\newcommand{\uq}{{\bf u}}
\newcommand{\Gq}{{\bf G}}

\newcommand{\Pq}{{\bf P}}

\newcommand{\Qq}{{\bf Q}}
\newcommand{\qq}{{\bf q}}

\newcommand{\yq}{{\bf y}}
\newcommand{\Wq}{{\bf W}}
\newcommand{\wq}{{\bf w}}

\newcommand{\Fq}{{\bf F}}

\begin{document}

\begin{frontmatter}

\title{Data-Driven Bilateral  Generalized Two-Dimensional Quaternion Principal Component Analysis with Application to Color Face Recognition}

\author[mysecondaryaddress,mymainaddress]{Mei-Xiang~Zhao}
\author[mythirdaryaddress,mysecondaryaddress]{Zhi-Gang~Jia\corref{mycorrespondingauthor}}
\cortext[mycorrespondingauthor]{Corresponding authors}
\ead{zhgjia@jsnu.edu.cn}
\author[mymainaddress2]{Dun-Wei~Gong\corref{mycorrespondingauthor}}
\ead{dwgong@vip.163.com}
\author[mymainaddress]{Yong~Zhang}
\address[mysecondaryaddress]{School of Mathematics and Statistics, Jiangsu Normal University, Xuzhou 221116, P. R.  China}

\address[mythirdaryaddress]{Research Institute of Mathematical Science, Jiangsu Normal University, Xuzhou 221116, P. R.  China}

\address[mymainaddress2]{School of Information Science and Technology,
Qingdao University of Science and Technology, Qingdao, Shandong, 266061,
P. R. China}

\address[mymainaddress]{School of Information and Control Engineering, 
China University of Mining and Technology,
Xuzhou 221116, P. R.  China}

\begin{abstract}
A new data-driven bilateral  generalized two-dimensional quaternion principal component analysis (BiG2DQPCA) is presented to extract the features of matrix samples from both row and column directions. This general framework directly works on the 2D color  images without vectorizing and well preserves the spatial and color information, which makes it   flexible to fit various real-world applications. A generalized ridge regression model of BiG2DQPCA is firstly proposed with orthogonality constrains on aimed features. Applying the deflation technique and  the framework of minorization-maximization,  a new quaternion optimization algorithm is proposed  to  compute the optimal features of BiG2DQPCA and   a closed-form solution is obtained at each  iteration. A new approach based on BiG2DQPCA is presented  for color face recognition and image reconstruction with a new data-driven weighting technique.  Sufficient numerical experiments are implemented on practical color face databases  and indicate the superiority of BiG2DQPCA over  the state-of-the-art methods in terms of recognition accuracies and rates of image reconstruction.
\end{abstract}

\begin{keyword}
 2DQPCA\sep   Color face recognition\sep   Quaternion optimization algorithm\sep  data-driven\sep  $L_p$ norm
\end{keyword}

\end{frontmatter}


\section{Introduction}\label{sec:introduction}
Two-dimensional principal component analysis \cite{yzfy04:2dpca} started a history of extracting features of grey image samples without converting them into high-dimensional vectors. With preserving spatial information, these  features were firstly taken from eigenvectors corresponding to large eigenvalues of the covariance matrix  generated by matrix samples from column direction. Then to obtain features preserving more geometric information, mathematics and scientific  scholars developed many powerful  variations of 2DPCA  by extracting features from row and column directions. However, there are still no 2DPCA-based model with $L_p$ norm to treat color image samples in the literature. In this paper,  we develop a new bilateral   generalized two-dimensional  quaternion principal  component analysis with $L_p$-norm (BiG2DQPCA) to compute features with  preserving their spacial and color information.

To recognize color  images,  it is very important to preserve the cross-channel correlation among red, green and blue channels based on the RGB color space.  This correlation is overlooked by  the traditional methods which   parallelly treat each  channel or  gather three color channels into a large matrix; see 
\cite{cxz20:LRQ,jjnz22} for more advantages of the quaternion representation.
Recently, many studies show that purely imaginary quaternion matrices are well adapted to color images by encoding the color channels into  three imaginary parts; and  the conventional principal component analysis (PCA) and 2DPCA  are  generalized  to quaternion domain, such as the quaternion PCA (QPCA) \cite{bisa03:QPCA}, two-dimensional QPCA (2DQPCA)  \cite{xyc15:CPCA},
the bilateral  2DQPCA \cite{scy11:bid2DQPCA}, the kernel QPCA (KQPCA) and the 2DKQPCA \cite{{cyjz17:KQPCA}}. Unfortunately, the non-commutativity of quaternion multiplication makes it impossible to straightforwardly  use classic fast algorithms to solve quaternionic (right) eigenproblems and optimization problems. This makes  the well developed quaternion variations of  PCA and 2DPCA far away from  practical application.  To overcome the computational difficulty, Jia \textit{et al.}\cite{jlz17:2DQPCA} proposed a new version of  2DQPCA, 
which constrains the features to simultaneously preserve the color and spatial information, and they computed these features in the row direction by the  well-known fast structure-preserving algorithms  \cite{jwl13:eigQ,jia2019}. 
  Xiao {\it et al.} further  presented a mathematically equivalent quaternion ridge regression model for 2DQPCA  in \cite{xz19:2DPCA-S} and added new sparsity constraints on the features.  Then Zhao \textit{et al.} \cite{zjg19:im-2dqpca}   improved  2DQPCA  by extracting  features from both row and column directions, which  saves the storage of compressed samples under feature subspaces and highly lifts the speeds of color face recognition and image reconstruction.  These works provide necessary fundamental models and methods for developing new variations of 2DQPCA and quaternion optimization algorithms. 

With beautiful geometry and   and powerful functionality,  $L_p$-norms are  playing an increasingly important role  in the advanced artificial intelligence models based on principal component analysis (PCA) or 2DPCA.   PCA  \cite{tp91}  has become a basic method  in the computer vision and pattern recognition.  To render PCA vulnerable to noises, the quadratic formulation of PCA is improved into $L_{1}$-norm or $L_{0}$-norm on the objection and constrain functions  in many sparse and robust models, such as  $L_{1}$-PCA \cite{kk05:L1pca}, $R_{1}$-PCA \cite{dz06:R1pca},  PCA-$L_{1}$ \cite{KN08:pcaL1}, and  a series of sparse PCA (SPCA) algorithms \cite{sl09,zh05,zht06:spca,sh08:spca,jnrs10:spca,ojht16:spca,rsbx18:spca,mzx12:rspca,pzlw19:rspca}.  
 Finally,  the generalized PCA (GPCA)\cite{lxz13:gpca,nk14:gpca} uses the $L_{p}$-norm   on the objective  and constraint functions, and extracts features of different geometric properties. 
In the same period,  a plenty of variations of 2DPCA were proposed in the literature, such as
2DPCA-$L_{1}$\cite{lpy10:2dpcaL1}, 2DPCA-$L_{1}$S \cite{mz12:RS2DPCA},  G2DPCA\cite{jw16:g2dpca} and many others \cite{cjcz19:R2DPCA, yj05:2dpca,zz05:2dpca}. Beside, 
 $L_p$-norms were successfully applied to develop new supervised models, such as  the well-known robust bilateral $L_p$-norm two-dimensional linear discriminant analysis \cite{lswd19}. 
 Motivated by these successful models, we  generalize 2DQPCA in \cite{jiaG2DQPCA}  to a new bilateral  variation with $L_p$ norm. 
 However, $L_p$-norms of quaternion vectors are still not well defined and let alone their applications.  So we need to start our work from  the basic definitions and  computational methods.

The contributions of this paper are listed  in three aspects.
\begin{enumerate}[(1)]
\item  A new bilateral  generalized 2DQPCA   with $L_{p}$-norm (BiG2DQPCA) is presented with extracting features from both row and column directions. 
This general framework directly works on the 2D color  images without vectorizing and well preserves the spatial and color information, which makes it   flexible to fit various real-world applications. 
\item A generalized ridge regression model of BiG2DQPCA is firstly proposed with adding the orthogonality constrains on aimed features. Applying a deflation technique and  the framework of minorization-maximization \cite{dk04},  a new quaternion optimization algorithm is proposed  to  compute the optimal features of BiG2DQPCA.   A closed-form solution is obtained at each step of iteration. 
\item  A new data-driven weighting technique is presented in the  BiG2DQPCA  approach  for color face recognition and image reconstruction. The extracted features are weighted by a learned optimal manner. 
Sufficient numerical experiments are implemented on practical color face databases to show  the data-driven learning of optimal weighting manner and the superiority of BiG2DQPCA over  the state-of-the-art methods.
\end{enumerate}

The paper is organized as follows. In Section \ref{s:preview}, we define the $L_p$-norm of quaternion vector and recall two models: 2DQPCA and  G2DPCA.   In Section \ref{s:BiG2DQPCA}, we present a new BiG2DQPCA with a new optimization algorithm.  In Section \ref{s:application}, we present  a BiG2DQPCA approach with   a new data-driven technique for color face recognition and image reconstruction.
In Section \ref{s:experiments}, we compare BiG2DQPCA  with  the advanced variations of 2DPCA in   numerical experiments on the GTFD, color FERET,  Faces95 and AR databases. Finally, a conclusion and future research topics are proposed in Section \ref{s:conclusion}.

\section{Preliminaries}\label{s:preview}
In this section, we  present  the  $L_p$-norm of quaternion vector and  recall two face recognition models (2DQPCA and G2DPCA).

\subsection{$L_p$-norm of quaternion vector}
\begin{figure}[!t]
  \centering
 \includegraphics[scale=0.6]{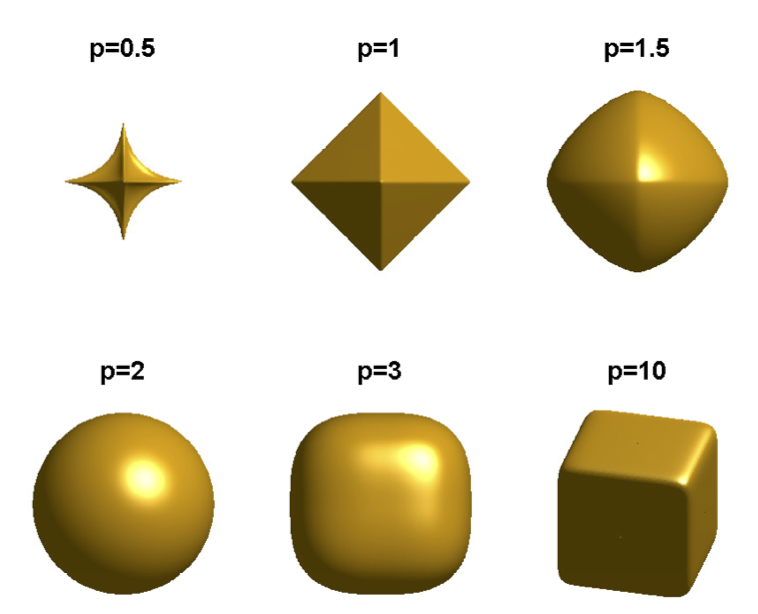}
\caption{ The set of vectors whose $L_p$ norms are less than or equal to one
}
\label{f:Lpnorm}
\end{figure}

The ring of real quaternions, denoted by
$$\mathbb{H}=\{\aq=a_0+a_1\iq+a_2\jq+a_3\kq~|~a_0,\cdots, a_3\in\mathbb{R}\},$$
where three imaginary values $\iq,~\jq,~\kq$ satisfy
$
   \iq^2=\jq^2=\kq^2=\iq\jq\kq=-1,
$
 was discovered by Sir William Rowan Hamilton (1805-1865) in 1843.
The absolute value  and the sign of  a quaternion number $\aq\in\H$ are defined by
 \begin{IEEEeqnarray}{rCl}\label{d:absign}
   |\aq|&=&(|a_0|^2+|a_1|^2+|a_2|^2+|a_3|^2)^{1/2},\IEEEyesnumber\IEEEyessubnumber\label{e:abs}\\
   {\rm sign}(\aq)&=&\left\{\begin{IEEEeqnarraybox}[\relax][c]{l's}
  \aq/|\aq|,&if $\aq\neq 0$,\\
  0,&if $\aq=0$.%
  \end{IEEEeqnarraybox}\right.\IEEEyessubnumber\label{e:sign}
 \end{IEEEeqnarray}
A quaternion vector $\wq\in\H^{n}$ is of the form
 \begin{equation*}\label{d:wq}
 \begin{aligned}
   \wq &=[\begin{array}{cccc}  \wq_{1}& \wq_{2} & \cdots &  \wq_{n}\end{array}]^T
   =w^{(0)}+w^{(1)}\iq+w^{(2)}\jq+w^{(3)}\kq
 \end{aligned}
\end{equation*}
where $\wq_{i} = w_{i}^{(0)}+w_{i}^{(1)}\iq+w_{i}^{(2)}\jq+w_{i}^{(3)}\kq$ denotes  the $i$th entry  of $\wq$ and $w^{(j)}\in\R^n$, $j=0,1,2,3$.
The sign and the absolute value of $  \wq$ are defined in the element-wise manner by
 \begin{IEEEeqnarray}{rCl}
   |\wq|&=&[\begin{array}{cccc} |\wq_{1}|&|\wq_{2}| & \cdots & |\wq_{n}|\end{array}]^T\IEEEyesnumber\IEEEyessubnumber\label{e:absv},\\
   {\rm sign}(\wq)&=&[\setlength{\arraycolsep}{0.2em} \begin{array}{cccc}  {\rm sign}(\wq_{1})& {\rm sign}(\wq_{2}) & \cdots &  {\rm sign}(\wq_{n})\end{array}\setlength{\arraycolsep}{5pt}]^T.\IEEEyessubnumber\label{e:signv}
 \end{IEEEeqnarray}
 And for zero quaternion vector $\vq=0$, we have ${\rm sign}(\vq)=0.$
Clearly,
  \begin{equation*}
  {\rm sign}(\wq)=
   \left[
                           \begin{array}{c}
                                                           \frac{w_{1}^{(0)}}{|\wq_{1}|} \\
                                                           \frac{w_{2}^{(0)}}{|\wq_{2}|} \\
                                                           \vdots \\
                                                           \frac{w_{n}^{(0)}}{|\wq_{n}|}\\
                                                         \end{array}
                                                       \right]+\left[
                                                                 \begin{array}{c}
                                                                   \frac{w_{1}^{(1)}}{|\wq_{1}|} \\
                                                                   \frac{w_{2}^{(1)}}{|\wq_{2}|} \\
                                                                   \vdots \\
                                                                   \frac{w_{n}^{(1)}}{|\wq_{n}|}  \\
                                                                 \end{array}
                                                               \right]\iq
                                                               +\left[
                                                                          \begin{array}{c}
                                                                            \frac{w_{1}^{(2)}}{|\wq_{1}|} \\
                                                                            \frac{w_{2}^{(2)}}{|\wq_{2}|} \\
                                                                            \vdots \\
                                                                            \frac{w_{n}^{(2)}}{|\wq_{n}|} \\
                                                                          \end{array}
                                                                        \right]\jq
                                                                        +\left[
                                                                                   \begin{array}{c}
                                                                                     \frac{w_{1}^{(3)}}{|\wq_{1}|} \\
                                                                                     \frac{w_{2}^{(3)}}{|\wq_{2}|} \\
                                                                                     \vdots \\
                                                                                     \frac{w_{n}^{(3)}}{|\wq_{n}|} \\
                                                                                   \end{array}
                                                                                 \right]\kq.
 \end{equation*}
Given a positive scalar $p$,   we define the $L_{p}$ norm of  quaternion vector $\wq$  by
\begin{equation}\label{e:Lpnorm}
  \|\wq\|_{p}=\left(\sum\limits_{i=1}^{n}|\wq_{i}|^{p}\right)^{\frac{1}{p}}.
\end{equation}
 If $p=2$, we get the $L_2$ norm of $\wq$.
 Note that the $L_{p}$ norms of quaternion vectors have different geometric properties.  See Figure \ref{f:Lpnorm} for illustration.

\subsection{2DQPCA}\label{S:2DCPAC}
2DQPCA was proposed  in \cite{jlz17:2DQPCA} for color face recognition.
 Suppose that an $m\times n$ quaternion matrix is in the form of $\Qq=Q_0+Q_1\iq+Q_2\jq+Q_3\kq$,   where  $Q_i \in \mathbb{R}^{m\times n}$, $i=0,1,2,3$. A  {\it pure quaternion matrix } is a matrix  whose elements are pure quaternions~$(Q_{0}=0)$~or zero.  In the RGB color space,  a color image of size  $m\times n$ can be stored by an $m\times n$  pure quaternion matrix $\Qq=[\qq_{ab}]_{m\times n}$  with each entry  $\qq_{ab}=r_{ab}\iq+g_{ab}\jq+b_{ab}\kq$  denoting one color pixel, where $r_{ab}$, $g_{ab}$ and $b_{ab}$ are red, green and blue channels \cite{pcd03}.

Suppose that there are $\ell$  training color image samples in total, denoted as $m\times n$ pure quaternion matrices,  $\Fq_1,\Fq_2,...,\Fq_\ell$,  and the mean  image of all the training color images can be defined as follows
${\bf\Psi}=\frac{1}{\ell}\sum\limits_{i=1}^{\ell}\Fq_i\in\H^{m\times n}.
$
Algorithm $\ref{m:2DQPCA}$ lists the procedure of 2DQPCA for image recognition.  2DQPCA based on quaternion models can preserve color and spatial information of face images, and its computational complexity of quaternion operations is similar to the computational complexity of real operations  of 2DPCA  (proposed in \cite{yzfy04:2dpca}).

\subsection{G2DPCA}\label{S:G2DPCA}
For treating real samples, the generalized two dimensional principal component analysis (G2DPCA) was proposed in \cite{jw16:g2dpca} by using $L_{p}$-norm in both the objective function and the constraint function.

 Suppose that  $F_1$, $F_2$, $\cdots$, $F_\ell$ $\in\mathbb{R}^{m\times n}$ are  $\ell$  real samples and they are centralized, that is,~$\sum_{i=1}^{\ell}F_{i}=0$.~G2DPCA solves the following optimization function
\begin{equation*}\label{G2DPCA}
  \mathop{{\rm max}}\limits_{w}\sum\limits_{i=1}^{l}\|F_{i}w\|_{s}^{s},~s.t.~\|w\|_{p}^{p}=1,
\end{equation*}
where $w\in\mathbb{R}^{n}$, $s\geq1$, and $p>0$.
In \cite{jw16:g2dpca}, an iterative algorithm is designed to solve the optimization problem of G2DPCA under the minorization-maximizatio (MM) framework \cite{dk04}, and a closed-form solution is obtained in each iteration. The solution of G2DPCA is guaranteed to be locally optimal.

  \begin{algorithm}
 \caption{\bf 2DQPCA: Two dimensional  quaternion principal component analysis}
 \label{m:2DQPCA}
   \begin{algorithmic}[1]
\STATE
  Compute the following {\it color image covariance matrix} of training samples
$
\Gq_t=\frac{1}{\ell}\sum_{i=1}^{\ell}(\Fq_i-{\bf\Psi})^*(\Fq_i-{\bf\Psi})\in\H^{n\times n},
$
where ${\bf\Psi}=\frac{1}{\ell}\sum_{i=1}^{\ell}\Fq_{i}$.
\STATE Compute the $k$ $(1\le k\le n)$ largest eigenvalues of $\Gq_t$ and their corresponding  eigenvectors (called eigenfaces), denoted as $(\lambda_1, \wq_1),$ $\ldots,$ $(\lambda_k, \wq_k)$.  Let the eigenface subspace
be $\Wq={\rm span}\{\wq_1,$ $\ldots,$ $\wq_k\}$.
\STATE  Compute the projections of $\ell$ training color face images in the subspace $\Wq$,
$\Pq_i=(\Fq_i-{\bf\Psi})\Wq \in\mathbb{H}^{m\times k},\ i=1,\cdots,\ell.
$
\STATE  For a given testing sample, $\Fq$,  compute its feature matrix, $\Pq=(\Fq-{\bf\Psi})\Wq$.
 Seek the nearest face image, $\Fq_i$ $(1\le i\le \ell)$, whose feature matrix satisfies that
$\|\Pq_i-\Pq\|=\min$.  $\Fq_i$ is output as the person to be recognized.
   \end{algorithmic}
 \end{algorithm}

\section{Bilateral  Generalized 2DQPCA}\label{s:BiG2DQPCA}
\noindent
In this section,  we present a bilateral  generalized 2DQPCA (BiG2DQPCA) with a generalized regression ridge model and a new quaternion optimization algorithm.

\subsection{Generalized ridge regression model}
Suppose that  $\Fq_1,\Fq_2,...,\Fq_\ell\in\mathbb{H}^{m\times n}$, are $\ell$ quaternion matrix samples and they are assumed to be mean-centered,  i.e., ${\bf\Psi}:=\frac{1}{\ell}\sum\limits_{i=1}^{\ell}\Fq_{i}=0$; otherwise, define $\Fq_{i}:=\Fq_{i}-{\bf\Psi}$ for $i=1,\cdots,\ell$.
The generalized ridge regression model of BiG2DQPCA  is 
\begin{equation}\label{e:big2dpcaQ}
  \begin{array}{cc}
    \left( \begin{array}{c}\widehat{\uq}_1,\cdots,\widehat{\uq}_{k_1},\\
                                          \widehat{\vq}_1,\cdots,\widehat{\vq}_{k_2}
                                          \end{array}\right)\!\!\!\!\!\!& =\!\!\!\!\!\!\mathop{{\rm arg~max}}\limits_{\uq_{j_1}\in\mathbb{H}^{m},~
\vq_{j_2}\in\mathbb{H}^{n}
}\sum\limits_{i=1}^{\ell}
\left(
\sum\limits_{j_1=1}^{k_1}\|\uq_{j_1}^*\Fq_{i}\|_{s}^{s}
+\sum\limits_{j_2=1}^{k_2}\|\Fq_{i}\vq_{j_2}\|_{s}^{s}
\right), \\
 &  {\text s. t.}~\|\uq_{j}\|_{p}=1,~\uq_{j}^{*}\uq_{i}=0 , i\neq j, i,j=1,2,\cdots,k_1, \\
  & ~~~~~~\|\vq_{j}\|_{p}=1,~\vq_{j}^{*}\vq_{i}=0 , i\neq j, i,j=1,2,\cdots,k_2, \\
  \end{array}
\end{equation}
where $s\geq 1,p>0$, and $k_1,~k_2$ are two given positive integers.
In this new model,  $L_{p}$-norms are used  in  both the objective function and the constraint function.  The constraints  $\|\uq_{j}\|_{p}\le 1$ and   $\|\vq_{j}\|_{p}\le 1$ are convex if $p\ge 1$ and non-convex if $0<p<1$.

Especially, if $s=p=2$, then BiG2DQPCA  reduces to a bilateral  two dimensional quaternion component analysis (Bi2DQPCA), which is not proposed in the literature.  Such Bi2DQPCA can be modelled in  a matrix form.
Define
$\Uq_{k_1}=[\uq_1,\cdots, \uq_{k_1}]$
and
$\Vq_{k_1}=[\vq_1,\cdots, \vq_{k_2}]$.
Applying the  Frobenius norm (F-norm),  Bi2DQPCA can be rewritten into
\begin{equation}\label{e:bi2DQPCAmatrix}
  \begin{array}{cc}
   &  (\widehat{\Uq}_{k_1},\widehat{\Vq}_{k_2}) =\mathop{{\rm arg~max}}\limits_{\Uq_{k_1},~\Vq_{k_2}}\sum\limits_{i=1}^{\ell}\left(\|\Uq_{k_1}^*\Fq_{i}\|_F^{2}+\|\Fq_{i}\Vq_{k_2}\|_F^{2}\right), \\
 &  {\text s. t.}~\Uq_{k_1}^{*}\Uq_{k_1}=I_{k_1}, ~~\Vq_{k_2}^{*}\Vq_{k_2}=I_{k_2}.
  \end{array}
\end{equation}
Moreover, from the constrains  that  the columns of $\Uq_{k_1}$ and $\Vq_{k_2}$ are orthonormal, we observe that
$$\|\Uq_{k_1}^*\Fq_{i}\|_F^{2}={\rm trace}(\Fq_{i}^*\Uq_{k_1}\Uq_{k_1}^*\Fq_{i})=\|\Uq_{k_1}\Uq_{k_1}^*\Fq_{i}\|_F^{2}$$
and
$$\|\Fq_{i}\Vq_{k_1}\|_F^{2}={\rm trace}(\Fq_{i}\Vq_{k_1}\Vq_{k_1}^*\Fq_{i}^*)=\|\Fq_{i}\Vq_{k_1}\Vq_{k_1}^*\|_F^{2}.$$
Then the matrix form of Bi2DQPCA is mathematically equivalent to
\begin{equation}\label{e:bi2DQPCAmatrixQRR}
\left.
  \begin{array}{cc}
    (\widehat{\Uq}_{k_1},\widehat{\Vq}_{k_2})\!\!\!\!\!&=\!\mathop{{\rm arg~min}}\limits_{\Uq_{k_1},~\Vq_{k_2}}\sum\limits_{i=1}^{\ell}\!\left(\|\Fq_{i}-\Uq_{k_1}\Uq_{k_1}^*\Fq_{i}\|_F^{2}   +\|\Fq_{i}-\Fq_{i}\Vq_{k_2}\Vq_{k_2}^*\|_F^{2}\right), \\
 &  {\text s. t.}~\Uq_{k_1}^{*}\Uq_{k_1}=I_{k_1}, ~~\Vq_{k_2}^{*}\Vq_{k_2}=I_{k_2}.~~~~~~~~~~~~
  \end{array}
\right.
\end{equation}

Remark that BiG2DQPCA of form \eqref{e:big2dpcaQ} is a novel and general model that covers  many well-known 2DPCA-based models.  Firstly, BiG2DQPCA is a generalization of   G2DPCA  \cite{jw16:g2dpca} from the real field to the quaternion skew-field and from one-directional feature extraction to two-directional feature extraction. Compared with G2DPCA , we firstly constrain the orthogonality of the projection vectors,  $\wq_{i},i = 1,2,\cdots,r$, in the ridge regression model \eqref{e:big2dpcaQ}.  Secondly, BiG2DQPCA  also generalizes 2DQPCA \cite{zjg19:im-2dqpca} by applying $L_{p}$-norm in both the objective function and the constraint function.  Thirdly, in the special case that $s=p=2$, Bi2DQPCA  of the matrix form \eqref{e:bi2DQPCAmatrixQRR} is a generalization of the QRR model  \cite{xz19:2DPCA-S} from one-directional model to bilateral  model.

\subsection{Quaternion optimization algorithm}
Now we present a new quaternion optimization algorithm of BiG2DQPCA by  solving the generalized  ridge regression  model \eqref{e:big2dpcaQ}. The pseudocode based on quaternion domain is given in Algorithm \ref{BiG2DQPCA}.  For simplicity of expression,  we define $\circledcirc$ by multiplying a real vector to a quaternion vector: Let $w\in\mathbb{R}^{n}$ and $\vq=v_{0}+v_{1}\iq+v_{2}\jq+v_{3}\kq\in\mathbb{H}^{n}$, then
\begin{equation*}\label{dotQ-mul}
  w\circledcirc\vq=(w\circ v_{0})+(w\circ v_{1})\iq+(w\circ v_{2})\jq+(w\circ v_{3})\kq,
\end{equation*}
where $\circ$ denotes the Hadamard product of two real vectors.
In  the practical computation we realize quaternion operations by the real structure-preserving method firstly proposed in \cite{jwl13:eigQ}.  The real representations of quaternion matrix $\Fq=F_0+F_1\iq+F_2\jq+F_3\kq\in\mathbb{H}^{m\times n}$ and quaternion vector $\wq\in\mathbb{H}^n$ are defined by 
\begin{equation*}\label{real-counterpart}
  \Fq^{(\Upsilon)}\equiv\left[
                 \begin{array}{rrrr}
                   F_0 & -F_1 & -F_2& -F_3 \\
                   F_1 & F_0 & -F_3 & F_2 \\
                   F_2 & F_3 & F_0 & -F_1 \\
                   F_3 & -F_2 & F_1 & F_0 \\
                 \end{array}
               \right]\in\R^{4m\times4n},
\end{equation*}
\begin{equation*}\label{d:real-vec}
  \wq^{(\gamma)}\!\equiv\!\left[
           \setlength{\arraycolsep}{0.2em}
            \begin{array}{llll}
              (w^{(0)} )^T&(w^{(1)})^T&(w^{(2)})^T&(w^{(3)})^T
            \end{array}
            \setlength{\arraycolsep}{5pt}
          \right]^T\!\in\!\R^{4n}.
\end{equation*}
The quaternion matrix-vector product $\yq=\Fq\wq$  is equivalent to the real matrix-vector product $\yq^{(\gamma)}=\Fq^{(\Upsilon)}\wq^{(\gamma)}$.

\begin{algorithm}
\caption{BiG2DQPCA: Bilateral  generalized two dimensional  quaternion principal component analysis with $L_p$ norm}
\label{BiG2DQPCA}
\begin{algorithmic}[1]
\REQUIRE {Training samples ~$\Fq_1, \Fq_2,\cdots, \Fq_{\ell}$, two positive integers  $k_1$ and $k_2$, and paraments $s\in [1,\infty), p\in(0,\infty]$. }
\ENSURE Optimal quaternion projection matrices $\Uq$ and $\Vq$, and weighted coefficient matrices $D^{left}$ and $D^{right}$.
\STATE Initialize $\Uq=[ \ ],~D^{left}=[ \ ],~\Vq=[\ ],~ D^{right}=[ \ ],~\Fq^{left}_{i}=\Fq^{right}_{i}=\Fq_{i}.$
\FOR{$t=1,2,\cdots,k_2$}
\STATE Initialize $k=0,~\delta=1$, arbitrary $\wq^0$ with $\parallel\wq^0\parallel_p=1$, and  $f^0=\sum\limits^{\ell}_{i=1}\|\Fq_{i}^{right}\wq^0\|_s^s.$
\WHILE{$\delta >10^{-4}$}
\STATE $\yq^{k}=\sum\limits^{\ell}_{i=1}(\Fq_{i}^{right})^{*}[|\widehat{\wq}^k|^{s-1}\circledcirc {\rm sign}(\widehat{\wq}^k)]$, where $\widehat{\wq}^k=\Fq_{i}^{right}\wq^k$.
\IF{$0<p<1$}
\STATE $\yq^k=|\wq^{0}|\circledcirc|\wq^k|^{1-p}\circledcirc\yq^k,$  $\wq^{k+1}={\yq^k}/{\parallel\yq^k\parallel_p}.$
\ELSIF{$p=1$}
\STATE $j=\mathop{\rm argmax}\limits_{i\in[1,n]}|\yq_{i}^{k}|,$
$ \wq_{i}^{k+1}=\left\{
\begin{aligned}
{\rm sign}(\yq_{j}^{k}), \ i=j,\\
0, \ i\neq j.\\
\end{aligned}
\right.$
\ELSIF{$1<p<\infty$}
\STATE $q=p/(p-1),$  $\yq^k=|\yq^k|^{q-1}\circledcirc{\rm sign}(\yq^k),$
 $\wq^{k+1}={\yq^k}/{\parallel\yq^k\parallel_p}.$
\ELSIF{$p=\infty$}
\STATE $\wq^{k+1}={\rm sign}(\yq^k).$
\ENDIF
\STATE $f^{k+1}=\sum\limits^{\ell}_{i=1}\|\Fq_{i}^{right}\wq^{k+1}\|_{s}^{s}.$
$\delta=|f^{k+1}-f^{k}|/|f^k|.$
\STATE $k\leftarrow k+1.$
\ENDWHILE
\STATE $\vq^{t}=\wq^{k}.$
\STATE Orthogonalize $\vq^{t}$ with the previous vectors $\vq_{1}$, $\cdots$, $\vq_{t-1}$ by quaternion QR decomposition.
\STATE $f^{t}=\sum\limits^{\ell}_{i=1}\|\Fq_{i}^{right}\vq^{t}\|_{s}^{s}.$
 $\Vq\leftarrow [\Vq,\vq^t]$,
 $D^{right}=[D^{right},f^t].$
\STATE $\Fq_i^{right} \leftarrow \Fq_i^{right}(\Iq-\Vq\Vq^{*}), i=1,2,\cdots,\ell.$
\ENDFOR
\STATE Run from steps $2$-$30$ again with replacing $k_2$ with $k_1$,  $\Fq_i\wq^k$ with $(\wq^k)^*\Fq_i$,  $\vq^t$ with $\uq^t$,  $\Vq$ with $\Uq$, $\Fq_i^{right}$ with $\Fq_i^{left}$,  $D^{right}$ with $D^{left}$,  and $\Fq_i^{right}(\Iq-\Vq\Vq^{*})$ with $(\Iq-\Uq\Uq^{*})\Fq_i^{left}$, respectively.   

\end{algorithmic}
\end{algorithm}

In Algorithm \ref{BiG2DQPCA},  the deflation technique is applied to provide  a proper searching subspace for computing the next principal component once the former ones have been obtained. Assume that the  first $k_1$ left projectors and the first $k_2$ right projectors have been computed, and define $\Uq_{k_1}=[\uq_{1},\uq_{2},\cdots,\uq_{k_1}]$ and $\Vq_{k_2}=[\vq_{1},\vq_{2},\cdots,\vq_{k_2}],$ 
where $1\leq k_1<m$ and $1\leq k_2<n$.
The next projectors $\uq_{k_1+1}$ and $\uq_{k_2+1}$ could be calculated similarly on the deflated samples
\begin{equation}\label{deflated}
  \Fq^{\textit{left}}_{i}=(\Iq-\Uq\Uq^{*}) \Fq_{i}, \quad
   \Fq^{\textit{right}}_{i}= \Fq_{i}(\Iq-\Vq\Vq^{*}),
\end{equation}
where $i= 1,2,\cdots,\ell.$
What is particularly noteworthy is that the additional  projection vector $\uq_{k_1+1}$ ( or $\vq_{k_2+1}$) obtained at each iteration must be orthonormalized against all previous  projection vectors by the modified Gram-Schmidt procedure in the quaternion domain.  
The reason is that once  we complete deflation in the known directions,  there is no information left in these directions and thus,  the new projection on the deflated samples should be zero.
We observe that after $(k_1+k_2)$ steps each sample is transformed into two deflated samples as in \eqref{deflated}. That is,  if  samples are projected into the first $k_1$ directions then
\begin{subequations}\label{orthonormal}
\begin{align}
&\uq_{j}^*(\Iq-\Uq\Uq^{*}) \Fq_{i}=0,~j=1,\cdots,k_1,\\
  &\Fq_{i}(\Iq-\Vq\Vq^{*})\vq_{j}=0,~j=1,\cdots,k_2.
 \end{align}
\end{subequations}
 If \eqref{orthonormal} does not hold, then the feature information can be lost because of the interference from other directions. From \eqref{orthonormal}, we also know that the next projector  can not be linearly represented by the prior projectors,  in other words, the newly computed projector is not  included in the subspace generated by the known projectors. So that we orthogonalize the computed projector to the known ones.

\section{Color face recognition with data-driven weighting}\label{s:application}
In this section,  we present a new BiG2DQPCA approach for color face recognition and reconstruction.
A new data-driven weighting technique is proposed. 

At first, we consider how to  weight the computed projections.   The  objective function values corresponding to the optimal  projection vectors $\uq_{1}, \cdots, \uq_{k_1}$ and $\vq_{1},\cdots, \vq_{k_2}$ are respectively  computed by 
$$f^{left}(k_1)=\sum\limits^{n}_{i=1}\left(
\sum\limits_{j_1=1}^{k_1}\|\uq_{j_1}^*\Fq_{i}\|_{s}^{s}
\right), \quad f^{right}(k_2)=\sum\limits^{n}_{i=1}\left(
\sum\limits_{j_2=1}^{k_2}\|\Fq_{i}\vq_{j_2}\|_{s}^{s}
\right).$$
Naturally, the increments
$$w^{left}(k_1)=f^{left}(k_1)-f^{left}(k_1-1),\quad w^{right}(k_2)=f^{right}(k_2)-f^{right}(k_2-1)$$
 characterize the contributions of  $\uq_{k_1}$ and $\vq_{k_2}$ to the objective function, and thus can be seen as their weighting factors, respectively. Then the joint weighting factor of the pair $(\uq_{k_1},\vq_{k_2})$ is  directly defined by $w(k_1,k_2)=w^{left}(k_1)+w^{right}(k_2)$. From now on, we denote the left eigenface subspace of  BiG2DQPCA by
 $\Uq=[\uq_1,$ $\ldots,$ $\uq_{k_1}]$ and the right eignface subspace by  $\Vq=[\vq_1,$ $\ldots,$ $\vq_{k_2}]$.
 Their weighting vectors are defined by $\Wq^{left}={\rm Diag}([w^{left}(1),$ $\ldots,$ $w^{left}(k_1)])$
 and $\Wq^{right}={\rm Diag}([w^{right}(1),$ $\ldots,$ $w^{right}(k_2)])$.

Under the assumption that all training samples are centralized, i.e., ${\bf{\Psi}}=0$,
the weighted projections of $\ell$ training face images on the subspaces $\Uq$ and  $\Vq$ are defined by
\begin{equation}\label{e:ps4fs2}
\Pq_i=(\Uq\Wq^{left})^*\Fq_i(\Vq\Wq^{right})\in\H^{k_1\times k_2},
\end{equation}
where $i=1,\cdots,\ell$.
Such $\Pq_i$ is called the {\it feature matrix} or {\it feature image} of the sample image $\Fq_i$.
In the experiments we use the 1-nearest neighbour (1NN) for the recognition with the feature matrix. In Algorithm $\ref{BiG2DQPCQ}$, we present a new BiG2DQPCA approach with  weighted projection for color image recognition.

 In the practical face recognition  tasks, we find the importances of features in  row and column directions  are  not equal to each other and the manner of weighting relies on the data.  
To  improve the efficiency of recognition, we replace the fixed weighting manner in \eqref{e:ps4fs2} by a new data-driven weighting technique.   

Let $f(\Wq^{left})$ and $f(\Wq^{right})$ be the functions of $\Wq^{left}$ and $\Wq^{right}$.    Four alternative weighting methods are given as follows.
\begin{enumerate}[(1)]
           \item Unweighted projection of face image $\Fq_i$ on the subspaces $\Uq$ and  $\Vq$ , i.e.,
           $$ \Pq_i=\Uq^*\Fq_i\Vq\in\H^{k_1\times k_2}.$$
           \item Left-weighted projection of face image $\Fq_i$ on the subspaces $\Uq$ and  $\Vq$ , i.e.,
           $$ \Pq_i=(\Uq f(\Wq^{left}))^*\Fq_i(\Vq)\in\H^{k_1\times k_2}.$$
           \item Right-weighted projection of face image $\Fq_i$ on the subspaces $\Uq$ and  $\Vq$ , i.e.,
           $$ \Pq_i=(\Uq)^*\Fq_i(\Vq f(\Wq^{right}))\in\H^{k_1\times k_2}.$$
           \item Bilateral  weighted projection of face image $\Fq_i$ on the subspaces $\Uq$ and  $\Vq$ , i.e.,
            $$ \Pq_i=(\Uq f(\Wq^{left}))^*\Fq_i(\Vq f(\Wq^{right}))\in\H^{k_1\times k_2}.$$
         \end{enumerate}

 \begin{algorithm}[H]
 \caption{\bf BiG2DQPCA with weighted projection for image recognition}
 \label{BiG2DQPCQ}
   \begin{algorithmic}[1]
\STATE
 For mean-centered  training samples $\Fq_{i},~i=1,2,\cdots,\ell$, compute the $k_1$ left and $k_2$ right $(1\le k_1\le m,~1\le k_2\le n)$ projection vectors and their corresponding weighting  vectors by Algorithm $\ref{BiG2DQPCA}$, denoted by $(\Uq,D^{left})$ and $(\Vq,D^{right})$.

\STATE Let  $\Wq^{left}={\rm Diag}(D^{left})$ and $\Wq^{right}={\rm Diag}(D^{right})$. Compute the projections of $\ell$ training color face images
\begin{equation*}\label{e:ps4fs22}\Pq_i=(\Uq\Wq^{left})^*\Fq_i(\Vq\Wq^{right}) \in\H^{k_1\times k_2},\ i=1,\cdots,\ell.\end{equation*}

\STATE For a given testing sample, $\Fq$,  compute its feature matrix, $\Pq=(\Uq\Wq^{left})^*\Fq(\Vq\Wq^{right})$.
 Seek the nearest face image, $\Fq_i$ $(1\le i\le \ell)$, whose feature matrix satisfies that
$i={\rm arg} \min\limits_{j}\|\Pq_j-\Pq\|$.  $\Fq_i$ is output as the person to be recognized.
   \end{algorithmic}
 \end{algorithm}

\begin{figure*}
           \centering
           \includegraphics[scale=0.6]{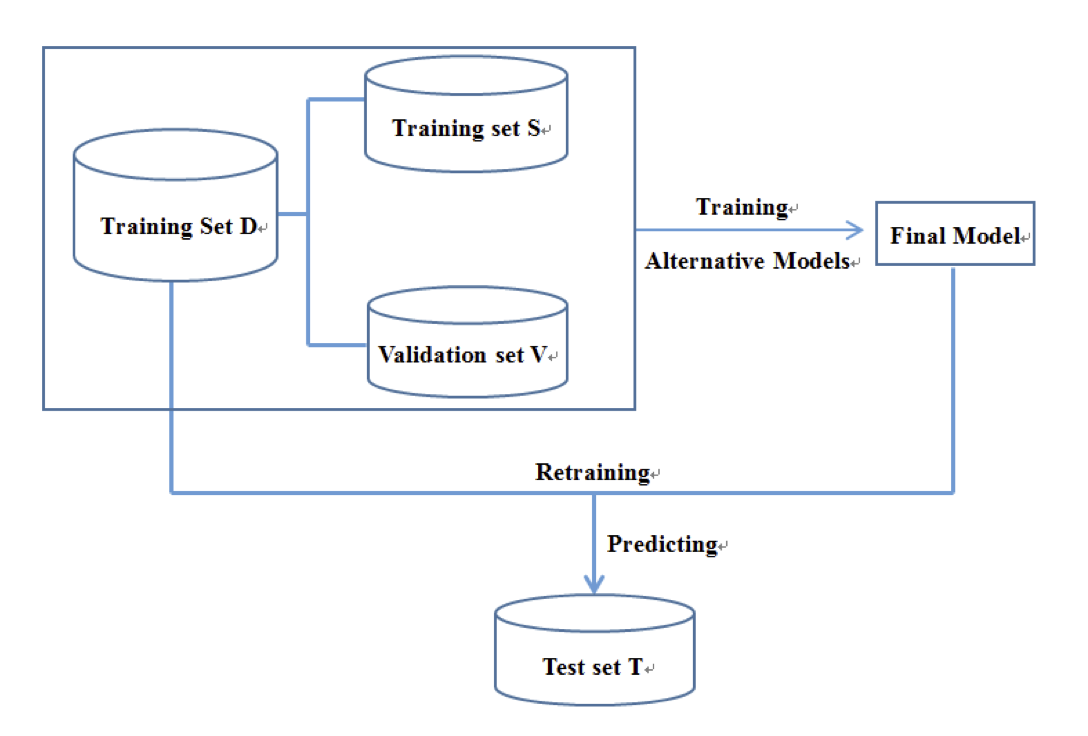}
           \caption{Flowchart of BiG2DQPCA  with data-driven weighting for face recognition}
           \label{flowchart}
\end{figure*}

Now the question becomes which  manner of weighting and what function $f$ should we choose.  The answer is to make a choice with relying on the property of the  obtained data. 
Let $D$ denote the training set of samples and  $T$ the test set.  We randomly select  two mutually exclusive sets from $D$: one  is seen as a new training set $S$ and another is used as the validation set $V$. 
Then we train BiG2DQPCA with four weighting formulae on $S$,  and test the recognition rate on  $V$.  In  general, we repeat the  random selection and the training-testing  process for several times, and then take the average values as the evaluation results, which help us to make a decision on the final weighting manner.  In our experiments, we repeat the whole process for three times and report the average results as the selection criteria. Once the weighting manner is decided, BiG2DQPCA is retrained with the original training set $D$.   
The flowchart of  BiG2DQPCA with data-driven weighting is illustrated in Figure $\ref{flowchart}$.

We close this section by applying BiG2DQPCA to color image reconstruction. Once  the subspace and the projections are computed by Algorithm \ref{BiG2DQPCA}, all we need is to project the projections of image samples back into the subspace.
The reconstruction process is described in Algorithm $\ref{alg:reco}$. Let $ \Fq_{i}^{rec}$ be the computed approximation to $\Fq_{i}$,  $i=1,\cdots,\ell$.  Then the ratio of image reconstruction is defined by 
\begin{equation}\label{e:ratio}Ratio=\frac{1}{\ell}\sum\limits_{i=1}^\ell\left(1-\frac{\|\Fq_{i}-\Fq_{i}^{rec}\|_F}{\|\Fq_{i}\|_F}\right).\end{equation}
\begin{algorithm}
 \caption{\bf BiG2DQPCA for color image reconstruction}
 \label{alg:reco}
   \begin{algorithmic}[1]
\STATE
For the mean-centered given training samples $\Fq_{i},~i=1,2,\cdots,\ell$, compute the left $k_1$ and right $k_2$ $(1\le k_1\le n, ~1\le k_2\le n)$ projection vectors by Algorithm $\ref{BiG2DQPCA}$, denoted as $(\Uq, \Wq^{left})$ and $(\Vq, \Wq^{right})$.

\STATE Add the average image of all training samples then
\begin{equation}
   \Fq_{i}^{rec}=\Uq(\Wq^{left})^{-1}\Pq_i(\Wq^{right})^{-1}\Vq^*.
\end{equation}
is output as the image to be reconstructed.
   \end{algorithmic}
 \end{algorithm}

\section{Experiments}\label{s:experiments}
\begin{figure}[!t]
           \centering
           \includegraphics[scale=0.65]{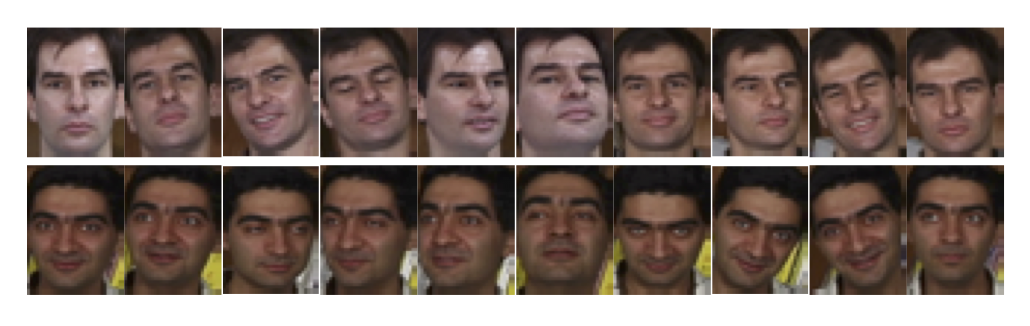}
           \caption{Sample color face images from the Georgia Tech face database. }
           \label{f:GTFD}
\end{figure}

\begin{figure}[!t]
  \centering
   \includegraphics[scale=0.65]{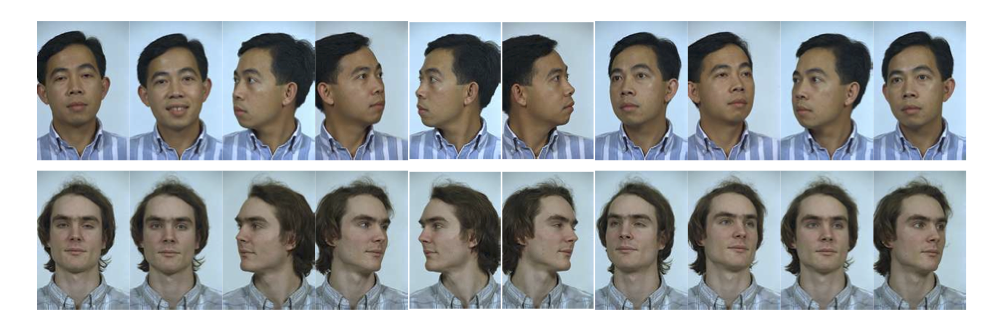}
\caption{Sample color face images from the color Feret database.}
\label{f:FERET}
\end{figure}

\begin{figure}[!t]
  \centering
 \includegraphics[scale=0.65]{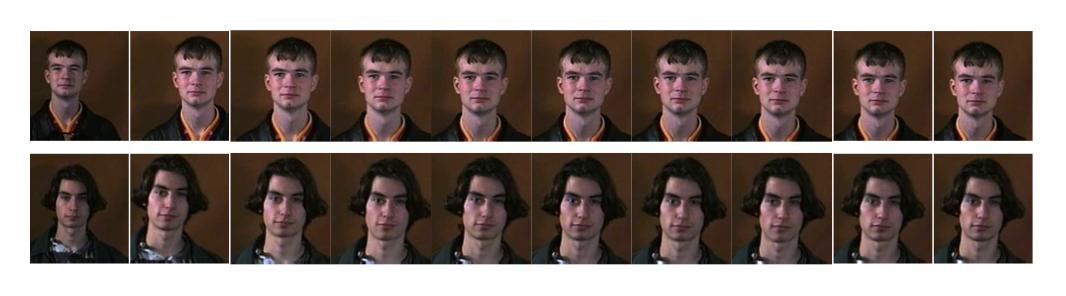}
\caption{Sample color face images from the faces95 database.}
\label{f:faces95}
\end{figure}

\begin{figure}[!t]
  \centering
 \includegraphics[scale=0.55]{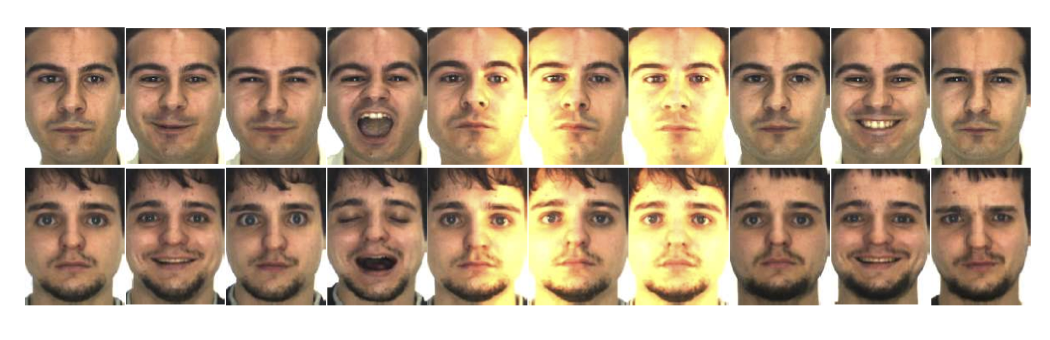}
\caption{Sample color face images from the AR database.}
\label{f:AR}
\end{figure}

\noindent
In this section , we  evaluate the proposed BiG2DQPCA 
 model on three face databases:
\begin{itemize}
\item GTFD---The Georgia Tech face database (GTFD)\footnote{\url{http://www.anefian.com/research/face\_reco.htm}.}  contains 750 images from 50 subjects, fifteen images per subject. It includes various pose faces with various expressions on cluttered backgrounds. All the images are manually  cropped, and then resized to $44 \times 33$ pixels. Some cropped images are shown in Figure  \ref{f:GTFD}.
\item ColorFeret---The color FERET database\footnote{\url{https://www.nist.gov/itl/iad/image-group/color-feret-database}.}  contains $1199$ persons, $14126$ color face images,  and each person has various numbers of face images with various backgrounds. The minimal number of face images for one person is $6$, and the maximal one is $44$.   The size of each cropped color face image  is $192\times 128$ pixels. Here we take $11$ images of $275$ individuals as an example. Some samples occluded with noise are shown in Figure $\ref{f:FERET}$.\\
\item Faces95---Faces95 database\footnote{\url{https://cswww.essex.ac.uk/mv/allfaces/faces95.html}.} contains $1440$ images photographed over a uniform background from $72$ subjects and $20$ images per subject. The size of each color face image is $200\times 180$ pixels. Some samples are displayed in Figure $\ref{f:faces95}$.

\item AR--- AR face database\footnote{\url{https://www2.ece.ohio-state.edu/~aleix/ARdatabase.html}.} contains frontal color face images of $126$ people recorded in two sessions. In each session, a neutral color face image is followed by images with different expressions and illumination, and images occluded by sunglasses and scarf (three images per condition). We employ a popular subset of AR containing $100$ subjects without occlusion, and hence, $1400$ images in total are used. Sample images from AR are given in Figure $\ref{f:AR}$.
\end{itemize}

The numerical experiments are to test the efficiencies of the methods in the task of image reconstruction and recognition with or without noise. We firstly compare the effects of four different weighting methods on the recognition results, then select the best weighting methods for face databases.  As mentioned in Section $\ref{s:application}$, the weighting vectors for $\Uq$ and $\Vq$ are denoted by $\Wq^{left}$ and $\Wq^{right}$, respectively. Here we consider four cases:  the projection matrices are not weighted (Unweighted), only  $\Uq$  is weighted by $\Wq^{left}$ on the left (Weighted Left), only $\Vq$ is weighted by $\Wq^{right}$ on the right (Weighted Right), and both $\Uq$ and $\Vq$ are weighted by $\Wq^{left}$ and $\Wq^{right}$ on the left and the right (Weightd Left and Right), respectively. For each database,  we randomly select 3 splits into 80\% training, 10\% validation and 10\% test images. The parameters are optimized on the validation set and tested on the test set.


\begin{example}[Weighting Manner Selection]
We firstly proceed to investigate the recognition performance on all four databases with different weighting methods. In this weight training process, the parameters $s$ and $p$ are fixed to $s=2$ and $p=2$.  The function $f$ of $\Wq^{left}$ and $\Wq^{right}$ is  set as $f(\Wq)=\Wq$.  The whole procedure is repeated three times and the average recognition rate is reported. Figures $\ref{cubeacc:GTFDfaces95}$ and $\ref{cubeacc:ARFeret}$ show the average recognition accuracy with four weighting manners on all the databases.  The numbers of the chosen eigenfaces in the column direction are shown in $X$ axes and the numbers of the chosen eigenfaces in the row direction are shown in $Y$ axes.
\begin{figure}
  \centering
  \includegraphics[width=0.99\textwidth,height=0.45\textwidth]{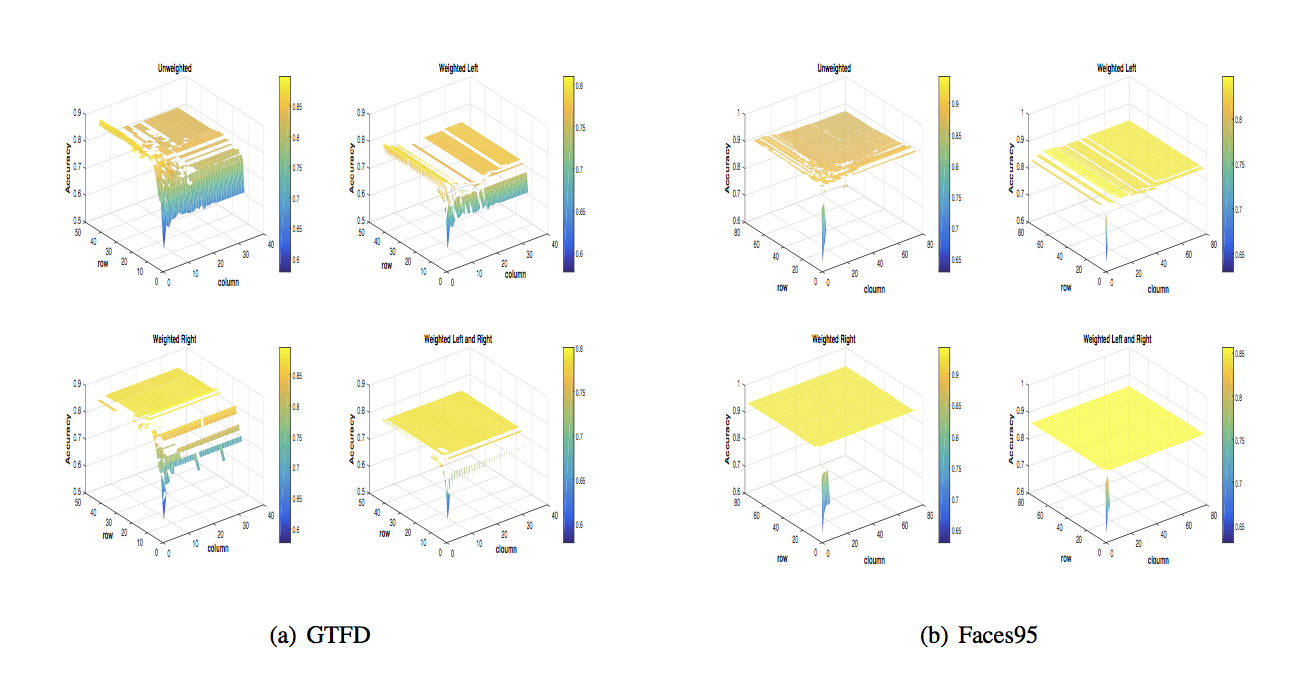}
\caption{Recognition accuracies of BiG2DQPCA with four different weighing method on the GTFD(left) and Faces95(right) validation sets.}
\label{cubeacc:GTFDfaces95}
\end{figure}

\begin{figure}
  \centering
  \includegraphics[width=0.99\textwidth,height=0.45\textwidth]{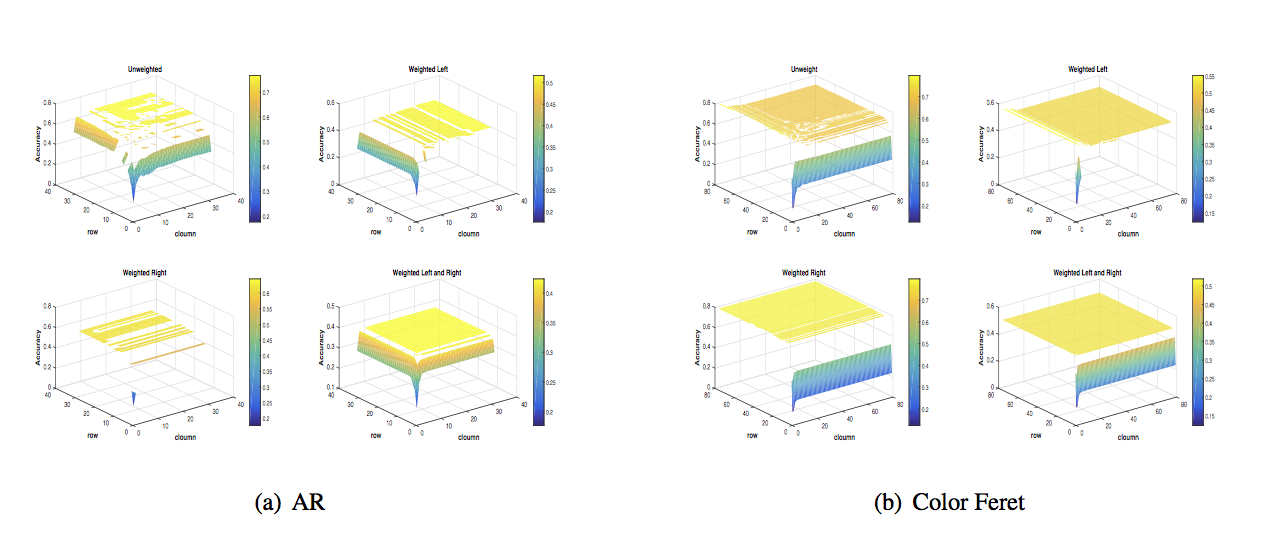}
\caption{Recognition accuracies of BiG2DQPCA with four different weighing methods on the AR(left) and Color Feret(right) validation sets.}
\label{cubeacc:ARFeret}
\end{figure}

As a special case, we select the same numbers of the eigenfaces in both row and column directions. And the recognition rates on validation set of four different databases are shown in Figure $\ref{30acc}$.  Here we notice that for AR database, the other three weighting methods seem not effectively compared with the unweighted case. So we try to take $f(\Wq)=1/\log(\Wq)$ and the results are shown in Figure $\ref{cubeacc:ARlog}$. The pictures display that for AR database, the projections with two-side weighting vectors achieve the highest classification accuracy if we select $f(\Wq)=1/\log(\Wq)$.
\begin{figure}[!t]
  \centering
  \includegraphics[width=0.99\textwidth,height=0.35\textwidth]{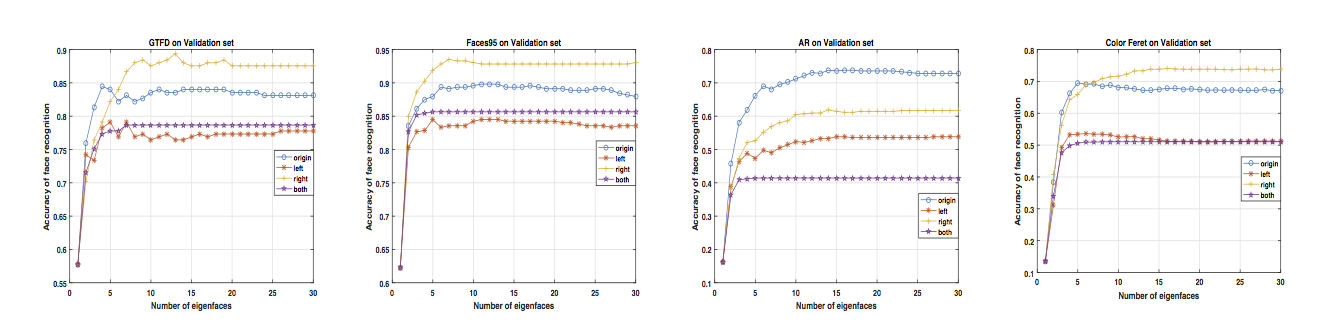}
\caption{Recognition accuracies of BiG2DQPCA with four different weighing method on the GTFD, Faces95, AR and Color Feret validation sets.  The numbers of the chosen eigenfaces are all from 1 to 30.}
\label{30acc}
\end{figure}

\begin{figure}
  \centering
  \includegraphics[width=0.99\textwidth,height=0.45\textwidth]{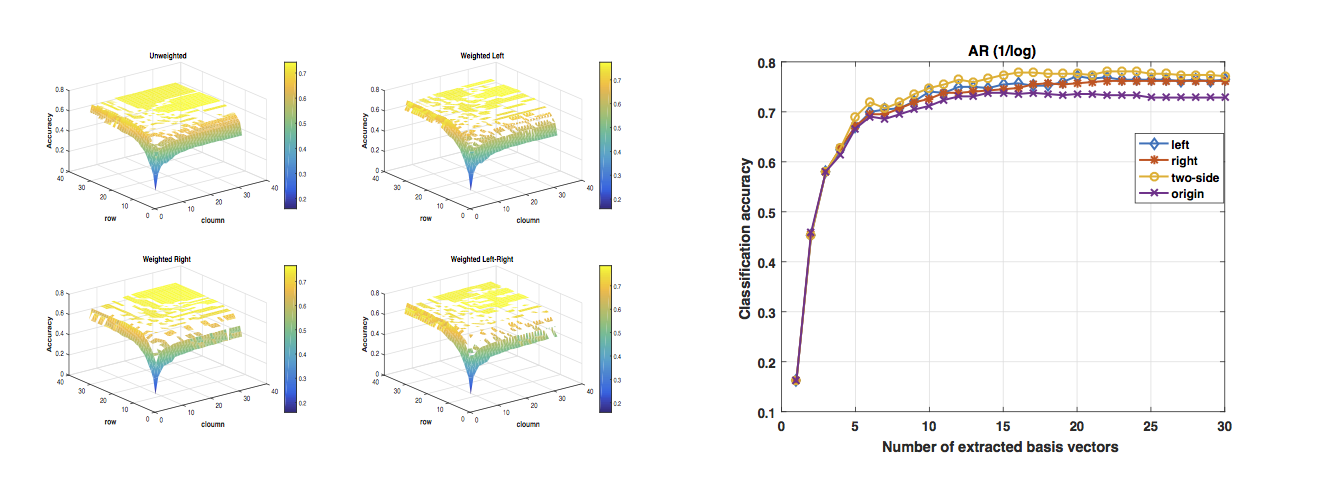}
\caption{Recognition accuracies of BiG2DQPCA on the AR validation set with $f(W)=1/\log(W)$.}
\label{cubeacc:ARlog}
\end{figure}

To further verify the reliability of the selected weighting method, we then test the rotated pictures of all databases. Some rotated samples are shown in Figure $\ref{rotatePic}$. Figure $\ref{transacc}$ displays that for the rotated GTFD, Faces95 and Color FERET databases, the projections with left weighting vectors achieve the highest recognition accuracy.
\begin{figure}[!t]
  \centering
  \includegraphics[width=0.99\textwidth,height=0.45\textwidth]{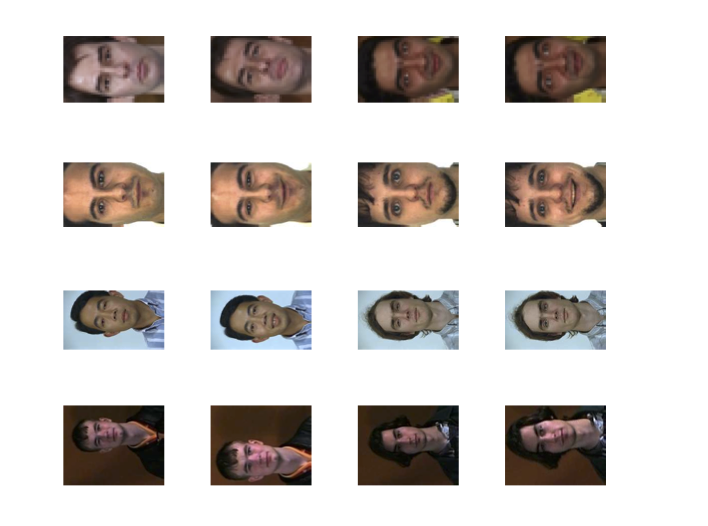}
\caption{Sample color face images from four rotated databases.}
\label{rotatePic}
\end{figure}
\begin{figure}
  \centering
   \includegraphics[width=0.99\textwidth,height=0.35\textwidth]{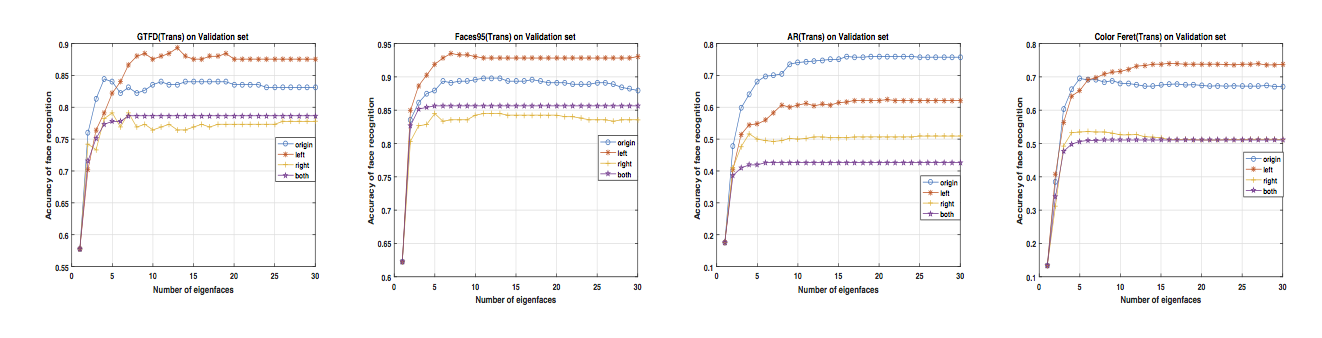}
\caption{Recognition accuracies of BiG2DQPCA with four different weighing methods on the rotated GTFD validation set.  The numbers of the chosen eigenfaces are from 1 to 30.}
\label{transacc}
\end{figure}

\end{example}

\begin{example}[Color Face Recognition]
In this example, we proceed to investigate the recognition performance of BiG2DQPCA (Algorithm $\ref{BiG2DQPCQ}$) and compare the results with several leading methods, including 2DPCA \cite{yzfy04:2dpca}, QPCA \cite{bisa03:QPCA}, 2DQPCA \cite{xyc15:CPCA}, G2DPCA \cite{jw16:g2dpca} and QRR \cite{xz19:2DPCA-S}. For GTFD, Faces95 and Color Feret databases, the weighting model chosen in the last experiment is the projection matrix $V$ weighted by $f(\Wq^{right})= \Wq^{right}$ on the right, that is, for a face image $\Fq$, its projection $\Pq$ equals to $\Uq^*\Fq(\Vq f(\Wq^{right}))$. For AR database, the weighting model is chosen as the projection matrices $\Uq$ and $\Vq$ are both weighted by $f(\Wq^{left})$ and $f(\Wq^{right})$ on left and right and the function $f$ is selected as  $f(\Wq)=1/\log(\Wq)$. For simplicity, let $w(s,p)$ denote BiG2DQPCA with $L_s$-norm and $L_p$-norm in all figures. Figure $\ref{ACCSP}$ shows the recognition accuracies of BiG2DQPCA with different parameter pairs $w(s,p)$ on four databases. The results indicate that the optimal parameter pair $(s,p)$ relies on the databases. Table $\ref{t:optpara}$ presents the best recognition rates on different databases and their corresponding optimal parameter pairs $(s,p)$.
\begin{figure}[!t]
  \centering
  \includegraphics[width=0.99\textwidth,height=0.35\textwidth]{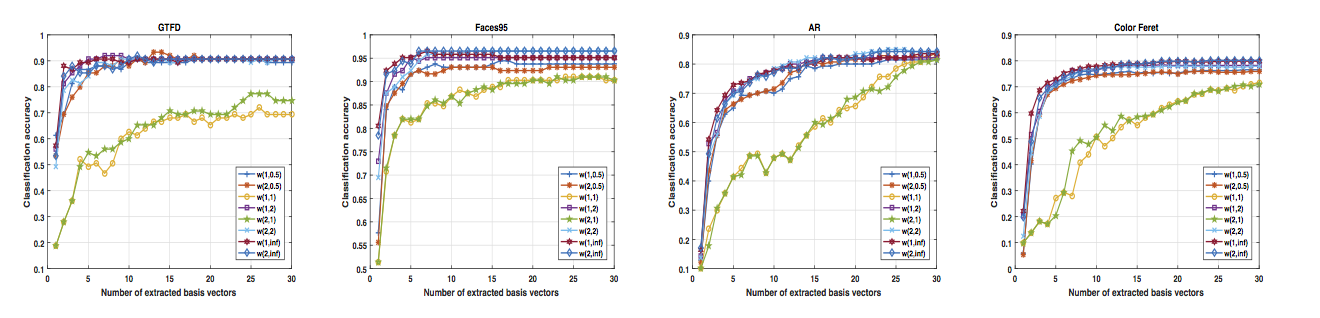}
\caption{Recognition accuracies of BiG2DQPCA with different $w(s,p)$ on the four different databases.}
\label{ACCSP}
\end{figure}

\begin{table}[h]
\caption{Optimal Parameter Pair $(s,p)$ on Four Databases}
\centering
\label{t:optpara}
\begin{tabular}{lll}
  \toprule
 Database &  Optimal parameters & Accuracy \\
 \midrule
 GTFD & s=2, p=0.5 & $93.33\%$ \\
 \midrule
 Faces95 & s=2, p=2 & $96.53\%$ \\
 \midrule
 Color Feret & s=2, p=$\inf$ & $80.20\%$ \\
 \midrule
 AR & s=2, p=2 & $85.00\%$ \\
\bottomrule
\end{tabular}
\end{table}

The following experiments compare the recognition accuracies of different algorithms on the color face images where the parameter pairs $(s,p)$ for BiG2DQPCA are set as those shown in Table $\ref{t:optpara}$. Figure $\ref{PCAlike}$ illustrates the comparison results.
\begin{figure}[!t]
  \centering
  \includegraphics[width=0.99\textwidth,height=0.35\textwidth]{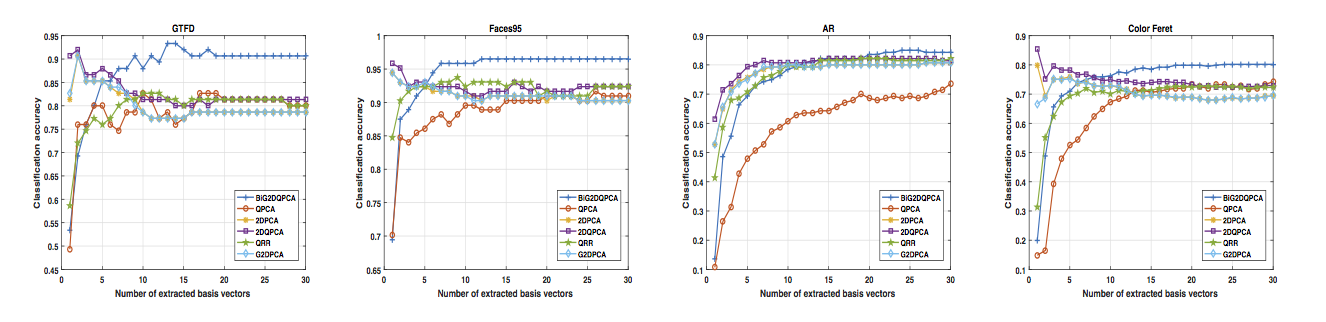}
\caption{Face recognition rate of PCA-like methods for the different databases.}
\label{PCAlike}
\end{figure}
\end{example}

\begin{example}[Color Image Reconstruction] In this example, we apply BiG2DQPCA (Algorithm $\ref{alg:reco}$) to reconstruct the color face images in training set from their projections.  The setting is as follows. For GTFD database, the number of eigenfaces changes from 1 to 44 and 1 to 33 in row space and column space, respectively. For AR database, the number of eigenfaces changes from 1 to 32 in both row direction and column direction. And for Color Feret and Faces95 databases,  the number of eigenfaces changes from 1 to 80 in both row space and column space. In Figure $\ref{ratio}$, we plot the ratios  of image reconstruction defined by \eqref{e:ratio} on four databases. These results indicate that the BiG2DQPCA is convenient to reconstruct color face images from projections, and can reconstruct original color face images with choosing all the eigenvectors to span the eigenface subspace.
\begin{figure}[!t]
  \centering
  \includegraphics[width=0.99\textwidth,height=0.35\textwidth]{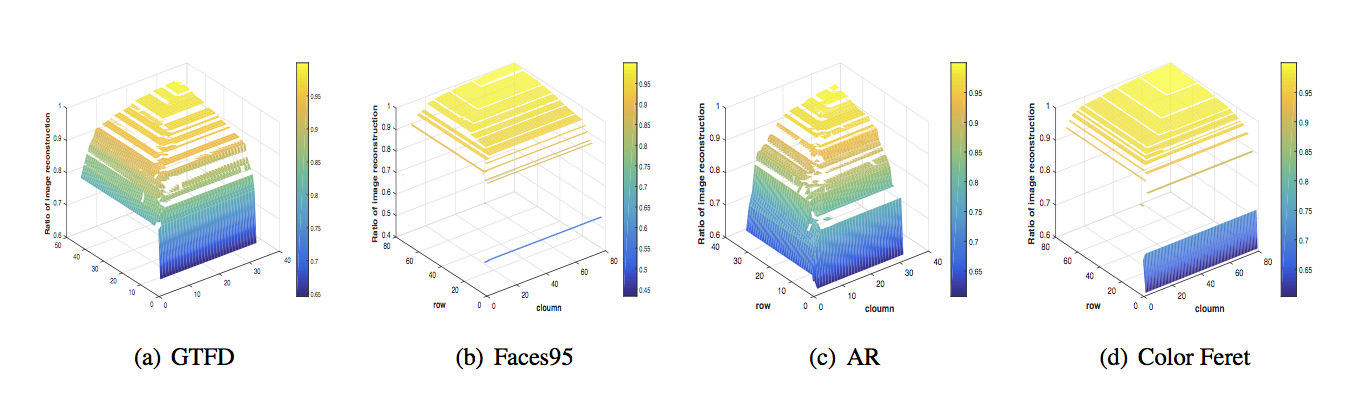}
\caption{Ratio of the reconstructed images over the original face images on the four databases.}
\label{ratio}
\end{figure}
\end{example}

\section{Conclusion and future works}\label{s:conclusion}
In this paper,  a novel BiG2DQPCA  is presented  for color face recognition and image reconstruction based on quaternion models. This general model generalizes famous 2DPCA-like methods, such as   2DPCA-$L_1$, Bi2DPCA and G2DPCA, from real field to quaternion ring,  and also generalizes 2DQPCA from $L_2$-norm constrain to $L_p$-norm constrain.   In addition,  BiG2DQPCA  extracts features of 2D color image samples from both row and column directions, and weights them by a new data-driven weighting technique.   The numerical experiments on the widely used face databases  (GTFD, Faces 95, AR, and color Feret) indicate the superiority of BiG2DQPCA over the state-of-the-art methods.

In the future, we will study a data-driven method of searching  optimal $L_s$-norm and $L_p$-norm in the BiG2DQPCA model and apply BiG2DQPCA to solve other color imaging problems, such as color image semantic segmentation and medical image processing.   We will also consider generalizing the two-dimensional linear discriminant analysis models in \cite{xcgz20:2DQS,yl05:2DLDA,nks06:2DLDA,lls08:2DLDA} to new variations with $L_p$ norm.

\section*{Acknowledgment}
This work was jointly supported by National Natural Science Foundation of China, Grant nos. 12171210, 12090011, 11771188, and 61773384; the Major
Projects of Universities in Jiangsu Province of China grants 21KJA110001; and the
Natural Science Foundation of Fujian Province of China grants 2020J05034.

\end{document}